# Robust Global Localization Using Clustered Particle Filtering

Adam Milstein, Javier Nicolás Sánchez, Evan Tang Williamson

Computer Science Department
Stanford University
Stanford, CA
{ahpmilst, jsanchez, etang} @cs.stanford.edu

**Abstract**

Global mobile robot localization is the problem of determining a robot's pose in an environment, using sensor data, when the starting position is unknown. A family of probabilistic algorithms known as Monte Carlo Localization (MCL) is currently among the most popular methods for solving this problem. MCL algorithms represent a robot's belief by a set of weighted samples, which approximate the posterior probability of where the robot is located by using a Bayesian formulation of the localization problem. This article presents an extension to the MCL algorithm, which addresses its problems when localizing in highly symmetrical environments; a situation where MCL is often unable to correctly track equally probable poses for the robot. The problem arises from the fact that sample sets in MCL often become impoverished, when samples are generated according to their posterior likelihood. Our approach incorporates the idea of clusters of samples and modifies the proposal distribution considering the probability mass of those clusters. Experimental results are presented that show that this new extension to the MCL algorithm successfully localizes in symmetric environments where ordinary MCL often fails.

## 1. Introduction

In mobile robotics, the task of navigation requires the ability for robots to identify where they are. Given a map of the environment and a starting pose (x-y position, orientation) in relation to the map, the task of localization is a tracking task. With unknown initial location, the task is known as global localization, in which a robot has to recover its pose from scratch [9,10]. The problem of localization is to compensate for sensor noise and errors in odometry readings.

One popular approach to robot localization is to use Kalman filters. Kalman filters are computationally efficient, but they require that the initial localization error be bounded---which makes them inapplicable to global localization problems. Additionally, Kalman filters assume linear-Gaussian measurement and motion dynamics. To overcome these limitations, a class of solutions was recently proposed that uses particle filters to represent the probability that the robot is in a particular location. This approach is commonly known as Monte Carlo Localization (MCL) [1].

In global localization, the robot starts off with no idea of where it is relative to its map. With a reasonably accurate map of the environment, MCL has been shown to be effective in many situations. However, MCL suffers an important limitation: When samples are generated according to their posterior probability (as is the case in MCL), they often too quickly converge to a single, high-likelihood pose. This might be undesirable in symmetric environments, where multiple distinct hypotheses have to be tracked for extended periods of time. MCL often converges to one single location too quickly, ignoring the possibility that the robot might be somewhere else. This problem leads to suboptimal behavior if there are two or more similarly likely poses. In symmetric environments, it is desirable to maintain a higher diversity of the samples, despite the fact that likelihood-weighted sampling will favor a single robot pose.

The approach we present in this article introduces the idea of clusters of particles and modifies the proposal distribution to take into account the probability of a cluster of similar poses. Each cluster is considered to be a hypothesis of where the robot might be located and is independently developed using the MCL algorithm. The update of the probability of each cluster is done using the same Bayesian formulation used in MCL, thus effectively leading to a particle filter that works at two levels, the particle level and the cluster level. While each cluster possesses a probability that represents the belief of the robot being at that location, the cluster with the highest probability would be used to determine the robot's location at that instant in time.

Experiments have been conducted with both simulated data as well as data obtained from a robot, using laser range finder data collected at multiple sites. The environments are highly symmetric and the corresponding datasets possess only a very small number of distinguishing features that allow for global localization. Thus, they are good testbeds for our proposed algorithm. Results show that the Cluster-MCL algorithm is able to successfully determine the position of the robot in these datasets, while ordinary MCL often fails.

## 2. Background

**Monte Carlo Localization and Bayes filter**

MCL is a recursive Bayes filter that estimates the posterior distribution of robot poses conditioned on sensor data. Central to the idea of Bayes filters is the assumption that the environment is Markovian, that is, past and future are conditionally independent given knowledge of the current state.

The key idea of Bayes filtering is to estimate a probability density over the state space conditioned on the data. This posterior is typically called the *belief* and is denoted by

$$Bel(x_t) = p(x_t \mid z_t, u_{t-1}, z_{t-1}, u_{t-2}, ..., z_0) \quad (1)$$

Here $x_t$ denotes the state at time $t$, $z_t$ is the *perceptual data* (such as laser range finder or sonar measurements) at time $t$, and $u_t$ is the *odometry data* (i.e. the information about the robot's motion) between time $t$-1 and time $t$.

Bayes filters estimate the belief *recursively*. The *initial* belief characterizes the *initial* knowledge about the system state, which in the case of global localization, corresponds to a *uniform distribution* over the state space as the initial pose is unknown.

To derive a recursive update equation, we observe that Expression (1) can be transformed by Bayes rule:

$$Bel(x_t) = \frac{p(z_t \mid x_t, u_{t-1}, ..., z_0) p(x_t \mid u_{t-1}, ..., z_0)}{p(z_t \mid u_{t-1}, ..., z_0)} \quad (2)$$

Because the denominator is a normalizer constant relative to the variable $x_t$ we can write equation (2) as

$$Bel(x_t) = \eta p(z_t \mid x_t, u_{t-1}, ..., z_0) p(x_t \mid u_{t-1}, ..., z_0) \quad (3)$$

where $\eta = p(z_t \mid u_{t-1}, ..., z_0)^{-1} \quad (4)$

As stated previously, Bayes filters make the Markov independence assumption. This assumption simplifies equation (3) to the following expression

$$Bel(x_t) = \eta p(z_t \mid x_t) p(x_t \mid u_{t-1}, ..., z_0) \quad (5)$$

The rightmost term in the previous equation can be expanded by integrating over the state at time $t$-1:

$$Bel(x_t) = \eta p(z_t \mid x_t) \int p(x_t \mid u_t, x_{t-1}, ..., z_0) p(x_{t-1} \mid u_{t-1}, ..., z_0) dx_{t-1} \quad (6)$$

And by application of the Markov assumption it can be simplified to

$$Bel(x_t) = \eta p(z_t \mid x_t) \int p(x_t \mid u_t, x_{t-1}) p(x_{t-1} \mid u_{t-1}, ..., z_0) dx_{t-1} \quad (7)$$

Defining $z^{t-1} = \{z_0, ..., z_{t-1}\}$ and $u^{t-1} = \{u_0, ..., u_{t-1}\}$ equation (7) can then be expressed as:

$$Bel(x_t) = \eta p(z_t \mid x_t) \int p(x_t \mid u_t, x_{t-1}) p(x_{t-1} \mid z^{t-1}, u^{t-1}) dx_{t-1} \quad (8)$$

It can be seen from equation (8) that the rightmost term is $Bel(x_{t-1})$. Therefore, this equation is recursive and is the update equation for Bayes filters. To calculate (8) one needs to know two conditional densities: the probability $p(x_t \mid u_t, x_{t-1})$, which is called the *motion model*, and the density $p(z_t \mid x_t)$, which is called the *sensor model*. The motion model is a probabilistic generalization of robot dynamics. The sensor model depends on the type of sensor being used and considers the noise that can appear in the sensor readings.

**Particle approximation**

If the state space is continuous, as is the case in mobile robot localization, implementing (8) is not trivial, particularly if one is concerned with efficiency. The idea of MCL is to represent the belief $Bel(x_t)$ by a set of $N$ weighted samples distributed according to $Bel(x_t)$:

$$Bel(x_t) \approx \{x_t^{[i]}, w_t^{[i]}\}_{i=1,...,N}$$

Here each $x_t^{[n]}$ is a *sample* of the random variable $x$, that is, a hypothesized state (pose). The non-negative numerical parameters, $w_t^{[n]}$, are called importance factors and they determine the importance of each sample. The set of samples thus define a discrete probability function that approximates the continuous belief $Bel(x_t)$.

In the case of global localization, the initial pose is unknown, thus the prior is uniform over the space of possible poses, and therefore each weight $w_t^{[n]} = 1/N$. Let $X_{t-1}$ be a set of particles representing the estimate $p(x_{t-1} \mid z^{t-1}, u^{t-1})$ at time $t-1$. The $t$-th particle set, $X_t$, is then obtained via the following sampling routine:

1. First, draw a random particle $x_{t-1}^{[n]}$ from $X_{t-1}$. By assumption, this particle is distributed according to $p(x_{t-1} \mid z^{t-1}, u^{t-1})$. Strictly speaking, this is only true as $N$ goes to infinity, but we ignore the bias in the finite case.
2. Next, draw a state $x_t^{[n]} \sim p(x_t \mid u_t, x_{t-1}^{[n]})$.
3. Finally, calculate the importance factor $w_t^{[n]} = p(z_t \mid x_t^{[n]})$ for this particle, and memorize the particle and its importance factor.

This routine is repeated $N$ times. The final set of particles, $X_t$ is obtained by randomly drawing (with replacement) $N$ memorized particles $x_t^{[n]}$ with probability proportional to the respective importance factor, $w_t^{[n]}$. The resulting set of particles is then an approximate representation for $Bel(x_t) = p(x_t \mid z^t, u^t)$. For a more detailed discussion on the implementation of MCL and examples see [6].

# 3. Global Localization using clustered particle filtering

We will now analyze how particle filters work and from that we will motivate our approach. To understand particle filters, it is worthwhile to analyze the specific choice of the importance factor. In general, the importance factor accounts for the "difference" between the target distribution and the proposal distribution. The target distribution is $p(x_t | z^t, u^t)$. The proposal distribution is given by $p(x_t | u_t, x_{t-1}) p(x_{t-1} | z^{t-1}, u^{t-1})$. This is the distribution of the samples values $x_t^{[n]}$ before the re-sampling step. The importance factor is calculated as follows:

$$w_t^{[n]} = \frac{\text{target distribution}}{\text{proposal distribution}}$$

$$w_t^{[n]} = \frac{p(x_t^{[n]} | z^t, u^t)}{p(x_t^{[n]} | u_t, x_{t-1}) p(x_{t-1} | z^{t-1}, u^{t-1})}$$

$$w_t^{[n]} = \frac{\eta p(z^t | x_t^{[n]}) p(x_t^{[n]} | u_t, x_{t-1}) p(x_{t-1} | z^{t-1}, u^{t-1})}{p(x_t^{[n]} | u_t, x_{t-1}) p(x_{t-1} | z^{t-1}, u^{t-1})}$$

$$w_t^{[n]} = \eta p(z_t | x_t^{[n]}) \quad (9)$$

The constant $\eta$ can easily be ignored, since the importance weights are normalized in the re-sampling step. This leaves the term $p(z_t | x_t^{[n]})$, which is the importance factor used in MCL.

Our analysis above suggests that a much broader range of functions may be used as proposal distributions. In particular, let $f_t(x_t)$ be a positive function over the state space. Then the following particle filter algorithm generates samples from a distribution $\propto f_t(x_t) p(x_t | z^t, u^t)$. Initially, samples are drawn from $f_0(x_0)$.

New sample sets are then calculated via the following procedure:
1. First, draw a random particle $x_{t-1}^{[n]}$ from $X_{t-1}$. By assumption, this particle is (asymptotically) distributed according to $f_{t-1}(x_{t-1}) p(x_{t-1} | z^{t-1}, u^{t-1})$ for very large $N$.
2. Next, draw a state $x_t^{[n]} \sim p(x_t | u_t, x_{t-1}^{[n]})$.

In this case the resulting importance factor is easily computed as:

$$w_t^{[n]} = \frac{\text{target distribution}}{\text{proposal distribution}}$$

$$w_t^{[n]} = \frac{f_t(x_t^{[n]}) p(x_t^{[n]} | z^t, u^t)}{f_{t-1}(x_{t-1}^{[n]}) p(x_t^{[n]} | u_t, x_{t-1}) p(x_{t-1} | z^{t-1}, u^{t-1})}$$

$$w_t^{[n]} = \frac{f_t(x_t^{[n]}) \eta p(z^t | x_t^{[n]}) p(x_t^{[n]} | u_t, x_{t-1}) p(x_{t-1} | z^{t-1}, u^{t-1})}{f_{t-1}(x_{t-1}^{[n]}) p(x_t^{[n]} | u_t, x_{t-1}) p(x_{t-1} | z^{t-1}, u^{t-1})}$$

$$w_t^{[n]} \propto p(z_t | x_t^{[n]}) \frac{f_t(x_t^{[n]})}{f_{t-1}(x_{t-1}^{[n]})} \quad (10)$$

The clustering particle filter proposed employs such a modified proposal distribution. In particular, each particle is associated with one out of $K$ clusters. We will use the function $c(x_t)$ to denote the cluster number. The function $f_t$ assigns to each particle in the same cluster the same value; but this value may differ among different clusters. Moreover, $f_t$ is such that the cumulative weight over all the particles in each cluster is the same for each cluster.

$$\sum_{x_t^{[n]} \in X_t : c(x_t^{[n]})=k} f(x_t^{[n]}) p(x_t^{[n]} | z^t, u^t)$$
$$= \sum_{x_t^{[n]} \in X_t : c(x_t^{[n]})=k'} f(x_t^{[n]}) p(x_t^{[n]} | z^t, u^t) \quad (11)$$

for $k \neq k'$.

From above we see that this is valid, however, we need to define $f(x_t^{[n]})$. Since these are equal for all $x$ such that $c(x) = k$, it is sufficient to define $f(x_t^{[n]}) = 1/B_{k,t}$ where $k = c(x_t^{[n]})$ and $B_{k,t}$ is the probability, at time $t$, that cluster $k$ contains the actual robot position. We can estimate the $B_{k,t}$ values using standard Bayes filters. Here, we use $k$ to represent the probability distribution over the clusters:

$$B_{k,t} = p(k_t | z^t, u^t)$$
$$= \eta \int p(z_t | k_t) p(k_t | u_t, k_{t-1}) p(k_{t-1} | z^{t-1}, u^{t-1}) dk_{t-1} \quad (12)$$

Since we use a finite number of samples to approximate the distribution, this becomes:

$$B_{k,t} = \eta \frac{\sum p(z_t | k_t) p(k_t | u_t, k_{t-1}) p(k_{t-1} | z^{t-1}, u^{t-1})}{n} \quad (13)$$

Now we note that, although the robot can move from one point to another, particles cannot change clusters. That is, each particle starts in one cluster and remains in that cluster. This being the case,

$$p(k_t | u_t, k_{t-1}) = \begin{cases} 0 & \text{if } k_t \neq k_{t-1} \\ 1 & \text{if } k_t = k_{t-1} \end{cases} \quad (14)$$

Therefore, $B_{k,t} \propto p(z_t | k_t) p(k_{t-1} | z^{t-1}, u^{t-1}) \quad (15)$

We also note that a cluster is composed of a set of points. Therefore, $p(z_t | k_t)$ is related to $p(z_t | x_t)$. In fact, the distribution of sensor readings for a cluster must be the sum of the distributions of sensor readings for all points in the cluster. That is:

$$p(z_t | k_t) \propto p(k_{t-1} | z^{t-1}, u^{t-1}) \sum_{x_t^{[n]} \in X_t : c(x_t^{[n]})=k} p(z_t | x_t^{[n]}) \quad (16)$$

Given equations (12) and (16) we can write

$$B_{k,t} = \gamma B_{k,t-1}^2 \sum_{x_t^{[n]} \in X_t : c(x_t^{[n]})=k} p(z_t | x_t^{[n]}) \quad (17)$$

where $\gamma$ is a normalization factor.

Having defined $f(x_t^{[n]}) = 1/B_{k,t}$, we maintain the condition stated in equation (11) by normalizing after each iteration. Therefore, we have shown that our modified proposal distribution is sound.

## 4. Cluster-MCL

**Algorithm:**

Based on the mathematical derivation above, we have implemented an extension to MCL, called Cluster-MCL. Cluster-MCL tracks multiple hypotheses organized in clusters. The first task is to identify probable clusters. By iterating several steps through ordinary MCL, with an initial uniform distribution of a large number of points, clusters develop in several locations. We then use a simple clustering algorithm to separate the points into different clusters. We match each point with a cluster based on the distance, in all three dimensions, between that point and the source point of the cluster. The initial probability of a cluster is based on the number of points it contains. There are more robust clustering algorithms, based on the EM algorithm; however, these methods rely on knowing the number of clusters a priori. Our method generates sets of clusters of arbitrary size. The drawback is that several clusters may be created in almost the same location. We solve this problem by occasionally checking for overlapping clusters and combining them. Once clusters are generated, we select the most probable ones and discard the others.

Each cluster is then independently evolved using ordinary MCL, thus points selected for a particular cluster can only be drawn from that cluster. The probability of each cluster is tracked by multiplying the prior probability of the cluster by the average of the likelihood of the points in that cluster. These probabilities are kept normalized and correspond to the $B_{k,t}$ values as defined above.

There is the problem that, if there is an error in the map in the initial location, there may be no cluster generated at the correct location. We solve this problem, and also the kidnapped robot problem, by taking advantage of the independence of our clusters. The kidnapped robot problem is where the robot is moved by an outside force after being localized. Since clusters do not interfere with each other, we can add a cluster in a new location without affecting our existing clusters. After a predetermined number of steps, we restart a new instance of global MCL with a higher convergence rate, with the purpose of finding the most likely cluster based on the current sensor data. Once global MCL has converged to a location, we check whether this new location overlaps an existing cluster. If it does not, we initialize it to have a small probability and begin tracking it. Otherwise, we discard it and repeat the process. By doing this, we remain open to consideration of a completely new location for the robot based on the current sensor data.

To limit the number of clusters from growing out of bounds and to remain computationally efficient, we limit the number of clusters to a maximum pre-defined value. Additionally, by keeping the number of clusters fixed at all points in time, we prevent a cluster from gaining a high probability by competing with only few other clusters, which would tend to prevent that cluster from being overtaken when there are many other clusters. When adding a new cluster, the least probable cluster is removed, in order to keep the size fixed.

The robot's estimate of its own location is based on the most likely cluster, and obtained by fitting a Gaussian through the corresponding particles.

## 5. Experimental Results

**Experimentation:**

The Cluster-MCL algorithm was implemented and tested in both simulated and real environments. In these tests, we compare the performance of our Cluster-MCL algorithm with that of ordinary MCL. In all cases, we found that Cluster-MCL performed as well as ordinary MCL, and in several cases where ordinary MCL failed, Cluster-MCL succeeded.

**Simulated Data.** For simulated environments, we generated two highly symmetrical maps to test on. Testing MCL and Cluster-MCL using these maps, we observed that Cluster-MCL correctly maintains all equally probable clusters, while ordinary MCL incorrectly and prematurely converges to a single cluster. In Figure 2, we display the results of Cluster-MCL using one of the maps, and we can clearly see that there are multiple distinct clusters. Notice that Cluster-MCL maintains a posterior belief comprised of four distinctive poses, in contrast to conventional MCL, whose outcome is shown in Figure 1. Moreover, the clusters in Figure 2 are all just about equally probable, as demonstrated by our observation of the constant trading off of which cluster is most probable. We obtained similar results on the second map, which was a simple rectangle.

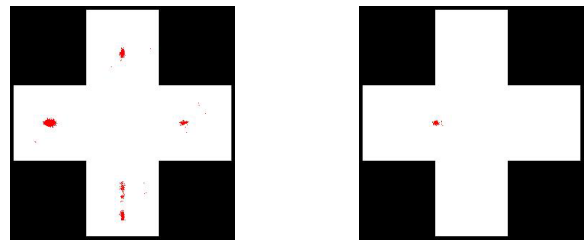

1(a)            1(b)

Figure 1: Global localization using ordinary MCL.

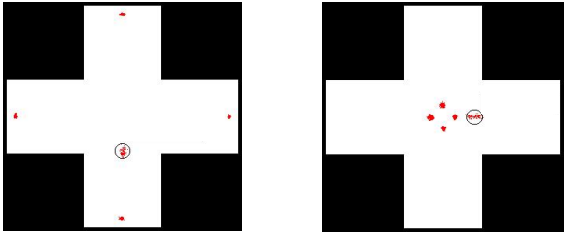

2(a)            2(b)

Figure 2: Global localization using Cluster-MCL. The extra cluster (circled) is a randomly drawn cluster, used to make Cluster-MCL robust to the kidnapped robot problem.

**Real Data.** To elucidate the workings of our algorithm in practice, additional tests were performed using data collected from two real world environments. Our first environment consisted of a long corridor in Wean Hall at Carnegie Mellon University, with equally spaced doors and few distinguishing features, thus providing an environment with some symmetry. Our second environment consists of a room in the Gates Building at Stanford University, with two entrances opposite each other, two benches symmetrically placed and a file cabinet in each corner of the room. The datasets in both locations were collected using a robot equipped with a laser range finder.

From these environments we collected nine datasets. From Wean Hall, we collected four datasets. In each dataset, the robot was given a different path with different features of the environment observed. Of the four cases, MCL was only able to correctly localize in three of them, while Cluster-MCL correctly identified the robot's position in all cases. In Figure 3, a comparison is given between MCL and Cluster-MCL on a particular dataset, number 3, from Wean Hall. On multiple executions over that particular dataset, ordinary MCL failed 100% of the time while Cluster-MCL had a 100% success rate. We show that ordinary MCL converges to the wrong location, while Cluster-MCL correctly identifies the robot's position.

In the Gates Building environment, five datasets on two different maps were collected. In all cases, Cluster-MCL performs at least as well as ordinary MCL. In four of the datasets, MCL and Cluster-MCL both correctly identify the robot's location. However, in the final dataset, MCL failed to consistently identify the correct location of the robot, while Cluster-MCL was able to localize to the correct position. The difference between the Wean Hall and Gates datasets is in the level of symmetry. To demonstrate the benefits of Cluster-MCL, we chose a more highly symmetrical environment in Gates and attempted to collect datasets, which had two possible localizations until the final segment of them. We ran MCL and Cluster-MCL several times on those datasets and the results show that MCL had 50% accuracy in determining the correct position, while Cluster-MCL had 100% accuracy.

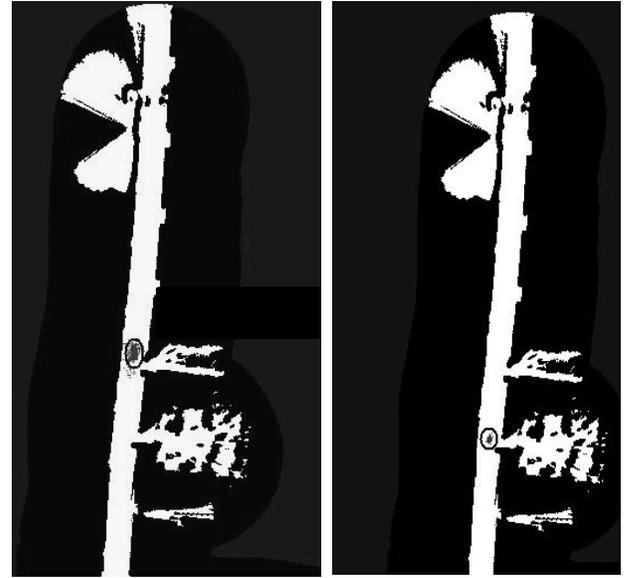

3(a)            3(b)

Figure 3: Results of MCL and Cluster-MCL on Wean Hall dataset 3. MCL converges to an incorrect cluster in 3(a), while Cluster-MCL converges to the correct location in 3(b).

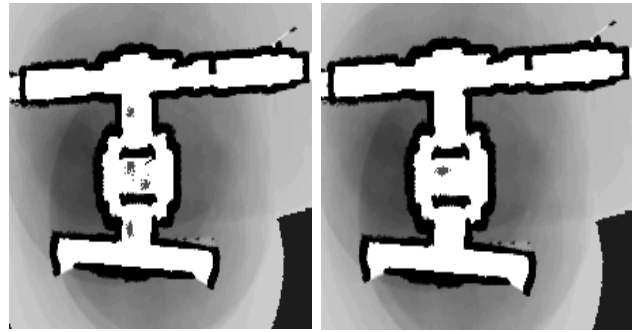

4(a)            4(b)

Figure 4: Results of MCL and Cluster-MCL on Gates data. Cluster-MCL tracks multiple possible clusters in 4(a) while ordinary MCL converges to a single, incorrect, cluster in 4(b).

## 6. Related Work

Related work in this area involves the use of multi-hypothesis Kalman filters to represent multiple beliefs. This however inherits Kalman filters limitations in that it requires noise to be Gaussian. A common solution to this problem is to perform low-dimension feature extraction, which ignores much of the information acquired by the robot's sensors [4,5]. Most of the work involving multi-hypothesis Kalman filters surrounds the tracking of multiple targets and feature detection, whereas we apply the concept of multiple hypotheses to represent our belief of the position of the robot.

Other improvements to MCL like, dual-MCL and Mixture MCL [6,11,12] attempt to improve the proposal distribution. Likewise, we attempt to improve the proposal distribution by way of tracking multiple hypotheses.

## 7. Conclusions and Future Work

**Conclusion**

In this paper we introduced a cluster-based extension to MCL localization. Ordinary MCL can fail if the map is symmetrical, however, we proposed a method that retains multiple hypotheses for where the robot is located, consistent with our sensor data. Our method involves clustering the points, and then tracking the clusters independently, so as to avoid discarding other possible locations in favor of the most probable cluster at the current time step. By considering the probability of multiple clusters over a longer time, we are able to get a more accurate idea of their likelihood. We have shown that this method is valid and that the additional information we take into account allows us to eliminate some of the bias from MCL and better approximate the true posterior. Our experiments show that Cluster-MCL performs at least as well as ordinary MCL on several real datasets, and in cases where MCL fails, Cluster-MCL still finds the correct location. Finally, we have shown that Cluster-MCL maintains all of the correct possible locations in symmetrical environments, while MCL converges to a single cluster.

**Future Work**

Future work might involve dynamically re-clustering the points on every time step in order to provide a true second-order MCL algorithm. We might also consider dropping clusters automatically when their probability drops below a certain threshold, instead of keeping a constant number of them. Since our algorithm retains less probable locations, it might be useful for a robot to consider less probable clusters when planning a motion. It might not be desirable for a robot to take an action that would be dangerous even if the corresponding cluster's likelihood is low. See [13] for an attempt to achieve this in the context of particle filtering.